\documentclass[10pt,twocolumn,letterpaper]{article}

\usepackage[table,xcdraw,dvipsnames]{xcolor}
\usepackage{iccv}              

\usepackage{changes}
\usepackage{pifont}
\usepackage{wrapfig}
\usepackage{etoolbox}
\usepackage{multirow}
\usepackage{mathtools}
\usepackage{algpseudocode}
\usepackage{xspace}
\usepackage{etoolbox}
\usepackage{graphicx}
\usepackage{epsfig} 
\usepackage{amsmath} 
\usepackage{amssymb}  
\usepackage{amsthm}
\DeclareMathAlphabet{\altmathcal}{OMS}{cmsy}{m}{n}
\usepackage{mathrsfs}
\usepackage{bm}
\usepackage{float}
\usepackage{bbm}
\usepackage{color}
\usepackage{lipsum}
\usepackage{placeins}
\usepackage{afterpage}
\usepackage{booktabs}  
\usepackage[tableposition=top]{caption}

\definecolor{StartBlue}{HTML}{2E8ED3}
\definecolor{EndGreen}{HTML}{2CBB00}
\definecolor{ReachPurple}{HTML}{7420CC}

\newcounter{FixCount}
\addtocounter{FixCount}{1}

\newcommand{\stkout}[1]{{\color{Plum}\ifmmode\text{\sout{\ensuremath{#1}}}\else\sout{#1}\fi}}

\newcommand{\norm}[1]{\left\Vert#1\right\Vert}

\providecommand{\R}{\ensuremath \mathbb{R}}

\providecommand{\opt}{\texttt{(Opt)}}
\providecommand{\splanningopt}{\texttt{(Splanning-Opt)}}

\providecommand{\T}{\mathcal{T}}

\providecommand{\E}{\mathbb{E}}

\providecommand{\T}{\ensuremath T}

\let\temp\phi
\let\phi\varphi
\let\varphi\temp

\newcommand{\pC}{c} 

\definecolor{tabfirst}{rgb}{1, 0.7, 0.7} 
\definecolor{tabsecond}{rgb}{1, 0.85, 0.7} 
\definecolor{tabthird}{rgb}{1, 1., 0.7} 

\definecolor{iccvblue}{rgb}{0.21,0.49,0.74}
\usepackage[pagebackref,breaklinks,colorlinks,allcolors=iccvblue]{hyperref}
\usepackage{svg}

\def\paperID{9022} 
\def\confName{ICCV}
\def\confYear{2025}

\title{These Magic Moments: Differentiable Uncertainty Quantification \\ of Radiance Field Models}

\author{Parker Ewen
\and
Hao Chen
\and
Seth Isaacson
\and
Joey Wilson
\and
Katherine A. Skinner
\and
Ram Vasudevan
}

\begin{document}
\twocolumn[{
    \renewcommand\twocolumn[1][]{#1}
    \maketitle
    \centering
    \includegraphics[width=1.0\linewidth,clip]{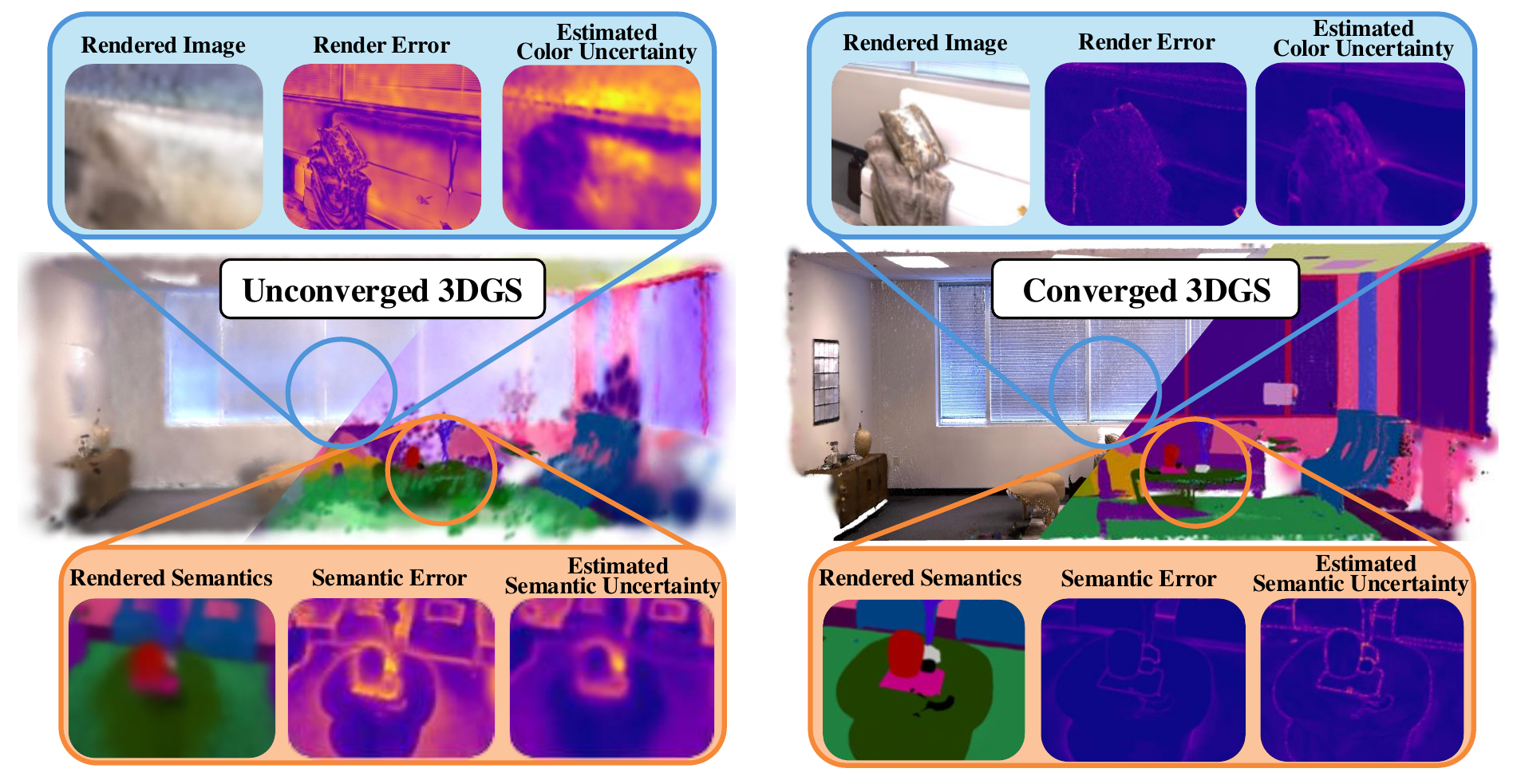}
    \captionof{figure}{Our proposed method quantifies uncertainty over rendered scenes generated through radiance field models by considering the differentiable higher-order moments of the rendering equation. 
    The proposed approach requires no additional training, generalizes for color, depth, and semantics, and may be used for both NeRF and 3DGS methods. 
    Shown is the variance, or second central moment, which is highly correlated to the rendering error for both color and semantics.
    }
    \label{fig:main} 
    \vspace{1em}
}]
\begin{abstract}
This paper introduces a novel approach to uncertainty quantification for radiance fields by leveraging higher-order moments of the rendering equation. 
Uncertainty quantification is crucial for downstream tasks including view planning and scene understanding, where safety and robustness are paramount. 
However, the high dimensionality and complexity of radiance fields pose significant challenges for uncertainty quantification, limiting the use of these uncertainty quantification methods in high-speed decision-making. 
We demonstrate that the probabilistic nature of the rendering process enables efficient and differentiable computation of higher-order moments for radiance field outputs, including color, depth, and semantic predictions. 
Our method outperforms existing radiance field uncertainty estimation techniques while offering a more direct, computationally efficient, and differentiable formulation without the need for post-processing.
Beyond uncertainty quantification, we also illustrate the utility of our approach in downstream applications such as next-best-view (NBV) selection and active ray sampling for neural radiance field training. 
Extensive experiments on synthetic and real-world scenes confirm the efficacy of our approach, which achieves state-of-the-art performance while maintaining simplicity.
\end{abstract}    
\section{Introduction} \label{sec:intro}

Radiance field models have become fundamental tools for novel view synthesis, enabling the generation of photo-realistic images from different perspectives based on a set of input views \cite{rabby2023beyondpixels}.
These models excel at learning continuous volumetric scene representations, facilitating high-quality, view-consistent rendering that can be applied to computer vision \cite{qin2024langsplat} and robotics \cite{michaux2024let, chen2024catnips} tasks.
Despite their widespread adoption, a notable challenge persists in quantifying the uncertainty within these models.
This limitation makes it difficult to determine when these models yield accurate scene representations, undermining their robustness and reliability, particularly for applications requiring high confidence in generated views.

Uncertainty quantification in radiance field models provides valuable insight into regions of the environment that the model represents poorly and areas where additional data could improve model accuracy.
This uncertainty can be leveraged for downstream applications such as active perception \cite{jin2023neu}, for safety-critical systems \cite{michaux2024let}, or for localization and mapping \cite{matsuki2024gaussian}.

Existing uncertainty quantification approaches for radiance field models typically attempt to model uncertainty in the model parameters directly, and approximate model output uncertainty in the form of rendered uncertainty \cite{sunderhauf2023density, goli2024bayes}.
However, the nonlinearity of the rendering function makes calculating pixel-wise probability densities from model parameter uncertainty mathematically intractable.
Existing approaches therefore rely on approximations or heuristics to transform model parameter uncertainty into rendering uncertainty, including training neural networks to estimate uncertainty directly \cite{jin2023neu, shen2022conditional}, approximating uncertainty over a 3D deformation field in a post-hoc manner \cite{goli2024bayes}, or relying on simplifying approximations \cite{jiang2023fisherrf}.
Furthermore, the complexity of these approximations often results in significantly increased computational costs. 

To address these challenges, this paper presents a formulation for quantifying uncertainty of radiance field model outputs directly.
We demonstrate that the probabilistic interpretation of the rendering equation from the underlying optical model enables efficient, differentiable computation of higher-order moments.
This approach shows equivalent or superior performance on downstream tasks compared to existing uncertainty quantification techniques, which also proving to be more expensive to compute.
The efficacy of this approach is demonstrated by first showing that the pixel-wise variance rendered using the proposed method has a higher correlation with rendering error than existing radiance field uncertainty quantification methods (Fig~\ref{fig:main}).
We further demonstrate that our uncertainty measure is a useful signal for two downstream tasks: next-best-view selection and active ray selection for neural radiance field model training.
In both tasks, the proposed approach outperforms baselines across all metrics.

In summary, the contributions of this paper are:
\begin{itemize}
    \item Derivation of pixel-wise higher-order moments for radiance field model rendering;
    \item Demonstration that the proposed pixel-wise variance strongly correlates to rendering error for color, depth, and semantic rendering;
    \item Application of pixel-wise uncertainty as a signal for next-best-view selection that outperforms existing state-of-the-art approaches;
    \item Application of pixel-wise uncertainty for active ray selection that results in higher quality neural radiance field models.
\end{itemize}

The remainder of the paper is organized as follows: 
Section \ref{sec:related} presents an overview of uncertainty quantification in radiance field methods from the literature.
Section \ref{sec:method} describes the probabilistic rendering process and the derivation of higher-order moments.
Section \ref{sec:experiments} demonstrates the efficacy of the proposed approach using three case studies: correlation with rendering error, next-best-view selection, and active ray sampling.
Finally, Section \ref{sec:conclusion} provides a discussion and concluding remarks.

\begin{figure*}
    \centering
    \includegraphics[width=0.85\linewidth]{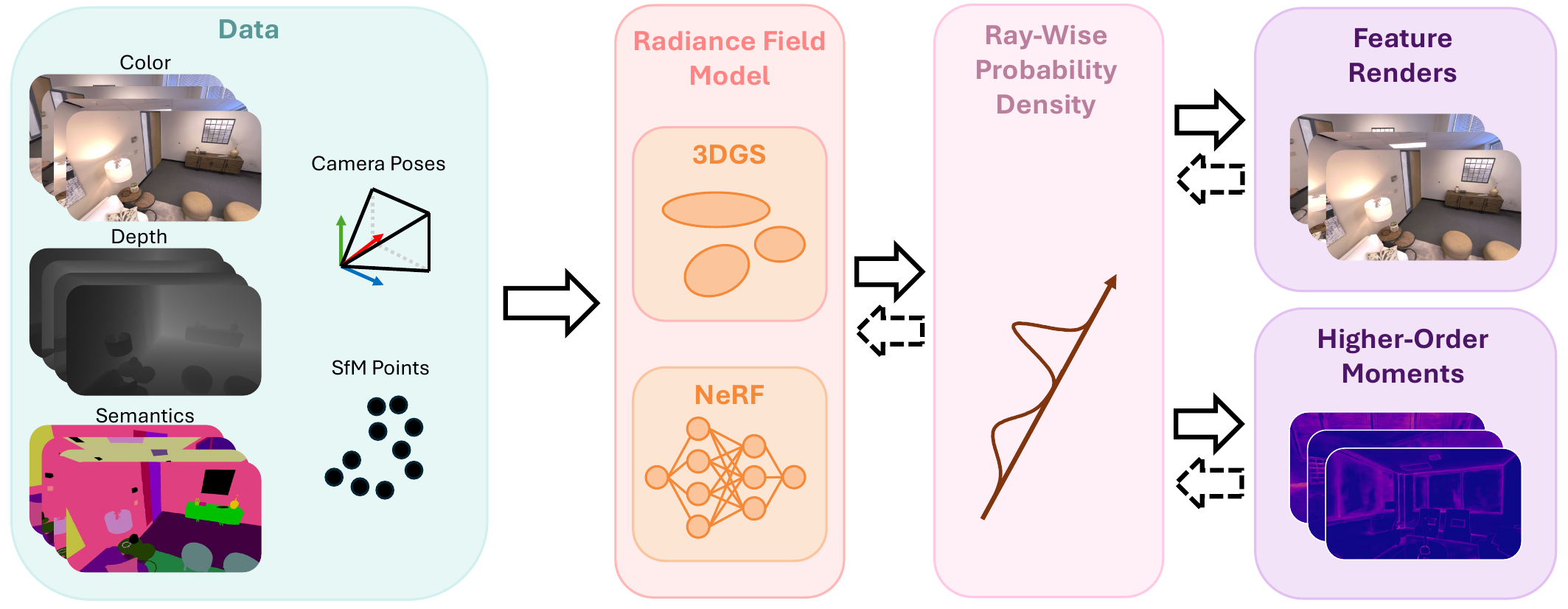}
    \caption{Images containing color, depth, semantics, or other features of interest, along with camera poses and an initial point cloud, are used to train a radiance field model depicted. The rendered features (i.e. the expected feature of each pixel) are computed in a differentiable manner. Our proposed approach demonstrates that the same probabilistic process used to compute these renders can be used to differentiably compute the higher-order moments. Solid arrows indicate operation flow and dotted arrows indicate gradient flow.
    }
    \label{fig:diagram}
\end{figure*}
\section{Related Works} \label{sec:related}
Uncertainty quantification plays a critical role in computer vision for enhancing model reliability and estimating prediction error \cite{szeliski1990bayesian}. 
This includes estimating both aleatoric uncertainty, which is uncertainty in the data resulting from measurement noise, and epistemic uncertainty, which is uncertainty in the model parameters \cite{amini2020deep}.
Ideally, uncertainty quantification for radiance field models should correlate with rendering error and be applicable to color, depth, and semantics.
Furthermore, computing the uncertainty should also be fast such that it can be used in downstream applications.

\subsection{Uncertainty In NeRF Models}
Radiance field models have seen a surge in popularity in recent years spurred by the development of neural radiance fields (NeRFs), which use a neural network to approximate the 5-dimensional radiance field function \cite{mildenhall2021nerf}.
NeRF-based methods are able to leverage existing uncertainty quantification methods for neural networks including network dropout \cite{gal2016dropout} and Bayesian neural networks (BNNs) \cite{kononenko1989bayesian, mullachery2018bayesian}, two well-established techniques used to quantify uncertainty using Monte Carlo and functional approximations, respectively.
While dropout has proven effective and offers a balance between computation cost and accuracy \cite{srivastava2013improving}, BNNs can be computationally intensive to train, may not generalize under domain shift, and may have trouble converging \cite{izmailov2021bayesian}.
Ensemble methods also approximate uncertainty, however, training a sufficient number of models is computationally intensive and inference time scales linearly with the number of models used \cite{sunderhauf2023density}, making ensemble methods unsuitable for many downstream tasks that require real-time operation.

Other prominent NeRF-based uncertainty quantification methods include S-NeRF \cite{shen2021stochastic} and CF-NeRF \cite{shen2022conditional}, which model uncertainty over the set of possible radiance field models via variational inference.
Evidential regression \cite{amini2020deep} has also shown promise for uncertainty quantification by training a network to output an additional variance parameter \cite{jin2023neu, shen2022conditional}, although fundamental concerns about the accuracy of evidential learning underpinning these approaches have been raised \cite{wu2024evidence, meinert2023unreasonable}.
Lastly, post-hoc approaches that approximate the volumetric uncertainty of a NeRF model without modifying the training process have also been proposed \cite{goli2024bayes}.
It was found that approximating uncertainty of the NeRF network parameters directly was infeasible due to high correlation between the network layers, and a spatial distortion field was found to accurately approximate geometric uncertainty.

Depth-Supervised NeRF representations frequently treat measured depth and depth computed by the NeRF as probability distributions, then consider each distributions' variance during the training process \cite{kangle2021dsnerf, rematas2022urf, isaacsonkung2023loner}.
These methods find that considering the variance in rendered depth improves convergence. 
Still, these methods do not use the variance to assess model quality and they have not been applied to color or semantics.

Often, NeRF-based uncertainty quantification methods require training multiple networks for both color and semantics, meaning uncertainty for each feature is estimated independently.
Furthermore, these NeRF-based methods are slow to estimate uncertainty as the uncertainty for each ray must be approximated via Monte Carlo estimation.

\subsection{Uncertainty In 3DGS Models}
Recently, explicit models using Gaussian basis functions to approximate the radiance field have gained popularity via 3D Gaussian Splatting (3DGS) \cite{kerbl20233d}.
These models trace their foundations to work in texture mapping \cite{heckbert1989fundamentals} and early computational volume rendering approaches \cite{zwicker2002ewa, max2005local}. 
They have shown state-of-the-art performance in novel view synthesis and scene representation \cite{chen2024survey} and uncertainty quantification methods for 3DGS models have recently emerged.
Similar to NeRF-based approaches, variational inference has also been used for uncertainty quantification in 3DGS models \cite{savant2024modeling}.
These variational inference methods again do not generalize to color, depth, and semantics, and require estimating uncertainty of each feature independently of the others.
Semantic uncertainty quantification for closed-set semantics has also been demonstrated for 3DGS models \cite{wilson2024modeling}.
This method uses the conjugate pair theorem \cite{ewen2022these} to track uncertainty over semantic estimates.

FisherRF \cite{jiang2023fisherrf}, a recent uncertainty quantification method for next-best-view selection, takes an information-theoretic approach to estimating uncertainty in the Gaussians' parameters, then renders a heuristic approximation of this uncertainty.
Unlike NeRF-based approaches, it was found that the parameters in the 3DGS model have minimal correlation \cite{jiang2023fisherrf}, and second-order Laplace estimates are used to capture this parametric uncertainty.
Notably, this information-theoretic uncertainty quantification using Fisher information requires a linear rendering model. 
As the Gaussian rasterization process is nonlinear, a Gauss-Newton approximation is made, which often over-approximates uncertainty when the number of Gaussian bases is not regularized.
Other methods use heuristics for uncertainty and leverage these heuristics for training and regularization \cite{kheradmand20243d, hu2024cg}.
While FisherRF has been shown to be useful for color and depth uncertainty quantification, the correlation with rendering error is low for unbounded scenes and it has not been extended to semantic uncertainty quantification.

In contrast to these approaches, this paper presents a general, unified approach for estimating the uncertainty in radiance field model outputs by considering the rendering equation as a probabilistic process.
The proposed approach works with both NeRF and 3DGS models and is able to render color, depth, and semantic uncertainty without any additional training.
Furthermore, we demonstrate that the uncertainty correlates strongly with the rendering error and outperforms state-of-the-art uncertainty quantification methods.
\section{Methodology} \label{sec:method}
This section presents an overview of the proposed uncertainty quantification approach for radiance field models.
First, a review of radiance fields and the associated rendering equation are presented in Section \ref{subsec:radiance-fields}.
Then, the rendering equation is re-examined through the lens of a probabilistic process using the local illumination optical model.
Finally, this probabilistic process is used to derive a means of rendering higher-order moments and variance using the NeRF and 3DGS radiance field models.

\subsection{Radiance Fields} \label{subsec:radiance-fields}
A radiance field models the image formation process by approximating a five-dimensional function that maps from a point and viewing direction to the density $\sigma: \R^3 \to \R$ and the color $\pC: \R^3 \times S^2 \to \R^3$ along the viewing ray.
Such radiance field models have also been extended to represent depth \cite{chung2024depth} and semantics \cite{qin2024langsplat}.
Under a radiance field model, an image is rendered by considering separately the ray that intersects the camera origin and each pixel.
Consider such a ray $r: \R \to \R^3$.
We define an additional quantity, the \textit{transmittance}, which defines the unnormalized probability that a particle traveling along ray $r$ from location $s=a$ to $s=b$ will experience a collision. 
This transmittance, denoted as $\T_a^b[r]$, is computed by
\begin{equation}\label{eq:app-transmittance}
    \T_a^b[r] = \exp\left(-\int_a^b \sigma(r(s))ds\right)
\end{equation}
From the transmittance, the quantity of interest $\rho: \R^3 \rightarrow \R$, representing either color, depth, or semantics, of the pixel corresponding to ray $r$ is computed as
\begin{equation}\label{eq:rendering-equation}
    I[\rho] = \int_{z_n}^{z_f} \rho(r(s))\sigma(r(s)) T_{z_n}^s[r]ds
\end{equation}
where $z_n, z_f$ are the near and far clipping planes.

\subsection{Uncertainty Quantification of Radiance Fields} \label{subsec:uncertainty-quant}

Early work in optical models considered volume rendering as a probabilistic process \cite{max1995optical, max2005local}.
The theory underpinning current radiance field approaches uses the so-called local illumination model, where a scene is comprised of groups of small, non-reflecting, opaque, glowing particles.
When rendering using this model, a ray is cast through this sea of particles and, when a ray collides with a particle at location $s$ along the ray, the color of the rendered pixel is the color emitted by the particle at location $s$.
An equivalent process may be defined for rendering depth and semantics.

These optical models treat the locations of these particles as uncertain.
The mass density of these particles is represented as $\sigma$, termed the density in the radiance field literature.
Ray collision is then, consequently, a random event, meaning the rendered quantity is a random variable \cite{max2005local}.

Under this probabilistic definition, $\sigma(r(s))$ represents the probability that a ray collides with a particle within the infinitesimal length $[s, ds]$ along the ray.
The transparency, or transmittance, $\T_0^s[r]$, then represents that probability that traveling from point $0$ to $s$ along the ray does not collide with a particle.
As these are independent events, the probability that a ray does not hit a particle between $0$ and $s$, and then hits one between distances $s$ and $ds$, is the multiplication of both probabilities, $\T_0^s[r] \sigma(r(s))$ \cite{tagliasacchi2022volume}.
Finally, $\T_0^s[r] \sigma(r(s))$ defines the probability density of particle collision along the ray taking the limit as $ds \rightarrow 0$.
Under these considerations, the rendering equation \eqref{eq:rendering-equation} represents the expected value of the random variable of interest, $\rho$, be it color, depth, or semantics \cite{max2005local}.

Extending this further, we propose computing the $j$-th moment for each pixel-wise random variable:
\begin{equation}\label{eq:rendering-moments}
    \E[\rho^j] = \int_{z_n}^{z_f} \rho(r(s))^j\sigma(r(s)) T_{z_n}^s[r]ds.
\end{equation}
The variance of each pixel is then computed using \eqref{eq:rendering-equation} and \eqref{eq:rendering-moments}:
\begin{equation} \label{eq:variance_render}
    \mathrm{Var}[\rho] = \E[\rho^2] - \E[\rho]^2.
\end{equation}

In computer vision, the terms ``variance" and ``uncertainty" are often taken as equivalent statements \cite{wilson2024convbki, wilson2024modeling, amini2020deep}.
We will refer to the variance as the quantified uncertainty for the remainder of this paper.

Similar formulas exist for other higher-order central moments if another another metric for uncertainty is desired.
Notably, the moment and central moment computations make no assumptions on the form of the random variable.
It is possible to use these  moments to fit a distribution to each pixel-wise random variable via moment fitting \cite{billingsley2013convergence}.

\begin{figure}
    \centering
    \includegraphics[width=\linewidth]{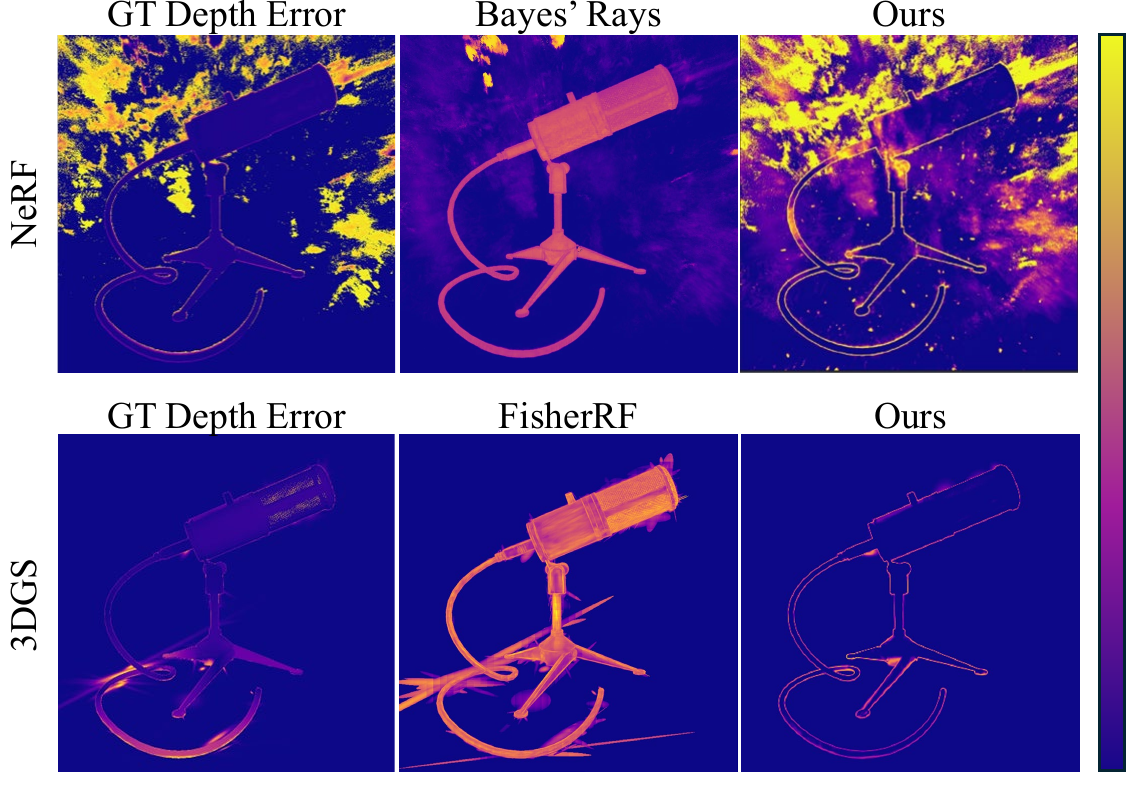}
    \caption{Uncertainty quantification for the depth estimates of a NeRF and 3DGS model. A NeRF model is trained and we compare the Bayes' Rays \cite{goli2024bayes} uncertainty estimate with our proposed variance estimate. Similarly, a 3DGS model is trained and we compare the FisherRF \cite{jiang2023fisherrf} uncertainty estimate with our proposed variance estimate. Unlike existing methods, our approach works with both NeRF and 3DGS models. The CF-NeRF \cite{yan2024cf} and 3DGS Ensemble \cite{sunderhauf2023density} methods are not shown as they require specialized models for uncertainty quantification.}
    \label{fig:depth_variance}
\end{figure}

\subsection{Uncertainty Computation}

Two prominent models have emerged for radiance field estimation.
The first is neural radiance fields, which use a feed-forward network to approximate features, $\rho$, and density, $\sigma$, of the radiance field function.
The second is 3D Gaussian Splatting models, which use a collection of Gaussian basis functions with differentiable components to represent the radiance field quantities.
Below we describe how to compute the moments for each model.
Figure \ref{fig:diagram} shows the algorithm for uncertainty quantification using a 3DGS model.

\subsubsection{Neural Radiance Fields}
Neural radiance fields jointly estimate the features, $\rho$, and density, $\sigma$, of the radiance field at discrete points along a ray cast from the camera center and through a pixel.
For volumetric rendering \eqref{eq:rendering-equation}, the ray is first divided into $N$ evenly spaced bins and one sample, $t_i$, is drawn uniformly from within each bin.
These samples are then used to approximate \eqref{eq:rendering-equation} via Monte Carlo integration such that, for each feature channel:
\begin{equation}
    \E[\rho] = \sum_{i=0}^{N} T_i c_i (1 - \exp(-\sigma_i \delta_i)),
\end{equation}
where $T_i = \exp(-\sum_{k=1}^{i-1} \sigma_k \delta_k)$, and $\delta_i = t_{i+1} - t_i$ is the distance between two adjacent samples.
To compute the moments of feature $\rho$, the same Monte Carlo Integration technique is applied to \eqref{eq:rendering-moments} such that:
\begin{equation} \label{eq:nerf_moments}
    \E[\rho^j] = \sum_{i=0}^{N} T_i c_i^j (1 - \exp(-\sigma_i \delta_i)).
\end{equation}

\begin{table*}[!t]
\centering
\footnotesize
\caption{Correlation coefficients for color variance and rendered color error. The CF-NeRF baseline did not converge on the TUM dataset and therefore no correlation coefficients are given. Presented is the time to render uncertainty for a single image. Cells are highlighted with red, orange, and yellow for for the first, second, and third best performance, respectively. Higher is better for the correlation coefficients.}
\begin{tabular}{l|ccc|c|ccc|c|ccc|c}
    \toprule
     \multirow{2}{*}{Method} & \multicolumn{4}{c|}{Blender} & \multicolumn{4}{c|}{Mip360} & \multicolumn{4}{c}{TUM Dataset} \\
     & $\tau_p$ & $\tau_s$ & $\tau_k$ & Time [ms] & $\tau_p$ & $\tau_s$ & $\tau_k$ & Time [ms] & $\tau_p$ & $\tau_s$ & $\tau_k$ & Time [ms]\\
     \midrule
     FisherRF \cite{jiang2023fisherrf} & \cellcolor{tabthird}0.580 & 0.215 & 0.137 & \cellcolor{tabsecond}$13679.3$ &\cellcolor{tabsecond}0.527 & -0.007 & -0.005 & \cellcolor{tabsecond}$156071.0$ & \cellcolor{tabsecond}0.482 & \cellcolor{tabthird}-0.071 & \cellcolor{tabthird}-0.046 & \cellcolor{tabsecond}$46769.2$ \\
     CF-NeRF \cite{yan2024cf} & 0.307 & \cellcolor{tabthird}0.612 & \cellcolor{tabthird}0.547 & $180718.0$ & 0.113 & \cellcolor{tabthird}0.187 & \cellcolor{tabthird}0.130 & $5329000.0$ & - & - & - & - \\
     3DGS Ensemble \cite{sunderhauf2023density} & \cellcolor{tabsecond}0.663 & \cellcolor{tabsecond}0.816 & \cellcolor{tabsecond}0.712 & \cellcolor{tabthird}$29.0$ & \cellcolor{tabthird}0.398 & \cellcolor{tabsecond}0.401 & \cellcolor{tabsecond}0.276 & \cellcolor{tabthird}$79.1$ & \cellcolor{tabthird}0.434 & \cellcolor{tabsecond}0.533 & \cellcolor{tabsecond}0.374 & \cellcolor{tabthird}$237.4$ \\
     Ours & \cellcolor{tabfirst}0.716 & \cellcolor{tabfirst}0.838 & \cellcolor{tabfirst}0.716 & \cellcolor{tabfirst}$2.0$ & \cellcolor{tabfirst}0.885 & \cellcolor{tabfirst}0.885 & \cellcolor{tabfirst}0.695 & \cellcolor{tabfirst}$6.8$  & \cellcolor{tabfirst}0.781 & \cellcolor{tabfirst}0.778 & \cellcolor{tabfirst}0.577 & \cellcolor{tabfirst}$20.4$ \\
     \bottomrule
\end{tabular}
\label{table:color_var}
\end{table*}
\begin{table*}[!t]
\footnotesize
\renewcommand{\arraystretch}{1.1}
\centering
\caption{Correlation coefficients for depth variance and rendered depth error. The CF-NeRF baseline did not converge on the TUM dataset and therefore no correlation coefficients are given. Presented is the time to render uncertainty for a single image. Cells are highlighted with red, orange, and yellow for for the first, second, and third best performance, respectively. Higher is better for the correlation coefficients.}
\begin{tabular}{l|ccc|c|ccc|c}
     \toprule
     \multirow{2}{*}{Method} & \multicolumn{4}{c|}{Blender} & \multicolumn{4}{c}{TUM Dataset} \\
     & $\tau_p$ & $\tau_s$ & $\tau_k$ & Time [ms] & $\tau_p$ & $\tau_s$ & $\tau_k$ & Time [ms]\\
     \midrule
     FisherRF \cite{jiang2023fisherrf} & \cellcolor{tabthird}0.572 & 0.328 & 0.238 & \cellcolor{tabthird}$13679.3$ & \cellcolor{tabthird}0.546 & \cellcolor{tabthird}0.575 & \cellcolor{tabthird}0.405 & $46769.2$ \\
     Bayes' Rays \cite{goli2024bayes} & \cellcolor{tabfirst} 0.758 & \cellcolor{tabthird}0.473 & \cellcolor{tabthird}0.364 & $45220.6$ & \cellcolor{tabsecond}0.596 & \cellcolor{tabsecond}0.638 & \cellcolor{tabsecond}0.468 & \cellcolor{tabthird}$46399.2$ \\
     CF-NeRF \cite{yan2024cf} & -0.086 & -0.305 & -0.346 & $180718.0$ & - & - & - & - \\
     3DGS Ensemble \cite{sunderhauf2023density} & 0.362 & \cellcolor{tabsecond}0.502 & \cellcolor{tabsecond}0.366 & \cellcolor{tabsecond}$29.0$ & 0.204 & 0.213 & 0.150 & \cellcolor{tabsecond}$237.4$ \\
     Ours & \cellcolor{tabsecond}0.718 & \cellcolor{tabfirst}0.792 & \cellcolor{tabfirst}0.700 & \cellcolor{tabfirst}$2.0$ & \cellcolor{tabfirst}0.671 & \cellcolor{tabfirst}0.685 & \cellcolor{tabfirst}0.519 & \cellcolor{tabfirst}$20.4$ \\
     \bottomrule
\end{tabular}
\label{table:depth_var}
\end{table*}
\begin{table}[!t]
\renewcommand{\arraystretch}{1.1}
\footnotesize
\centering
\caption{Correlation coefficients for semantic variance and rendered semantic error. Ground truth semantics are computed by processing the training RGB images through the CLIP-LSeg model.}
\begin{tabular}{l|ccc}
     \toprule
     \multirow{2}{*}{Method} & \multicolumn{3}{c}{Mip 360} \\
     & $\tau_p$ & $\tau_s$ & $\tau_k$ \\
     \midrule
     Ours & 0.180 & 0.200 & 0.135  \\
     \bottomrule
\end{tabular}
\label{table:semantic_var}
\end{table}

\subsubsection{3D Gaussian Splatting}
Recently, 3D Gaussian Splatting (3DGS) models have shown fast and efficient performance in representing radiance fields using 3D differentiable Gaussian basis functions.
Each 3D Gaussian consists of an opacity, $\alpha$, mean and covariance of the Gaussian, $\mu$ and $\Sigma$, as well as a color or semantics, here represented as the learned feature $\rho$ \cite[Section 2.3]{kerbl20233d}.
We combine all parameters together such that $\boldsymbol{w} = \{\alpha_i, \mu_i, \Sigma_i, \rho_i\}_{i=1}^M$, where $M$ is the number of Gaussians in the splatting model.

Unlike neural radiance fields, Monte Carlo Integration along rays is not required for Gaussian Splatting as the 3D Gaussians are projected onto the image plane and subsequent pixel colors, depths, and semantics are computed.
Given a camera pose $(\boldsymbol{x}, \boldsymbol{d}) \in SE(3)$ consisting of position $\boldsymbol{x} \in \R^3$ and viewing direction $\boldsymbol{d} \in SO(3)$, the rendering function \eqref{eq:rendering-equation} for a given feature channel of a pixel, $\rho$, in the image plane is
\begin{align} \label{eq:gaussian-render}
    \E[\rho] = \sum_{i\in \mathcal{I}} \rho_i \Tilde{\alpha}_i(\boldsymbol{x}, \boldsymbol{d}) \prod_{k<i}(1-\Tilde{\alpha}_k(\boldsymbol{x}, \boldsymbol{d}))
\end{align}
where $\rho_i$ is either the color, depth, or semantic embedding of the $i$-th ordered Gaussian and
\begin{align} \label{eq:alpha_comp}\Tilde{\alpha}_i(\boldsymbol{x}, \boldsymbol{d}) &= \alpha_i \cdot \exp\left( -\frac{1}{2}\norm{\boldsymbol{x} - \boldsymbol{\mu}_i(\boldsymbol{d})}_{\Sigma_i^{-1}(\boldsymbol{d})}^2 \right)
\end{align}
is the pixel-wise opacity of the $i$-th ordered Gaussian in set $\mathcal{I}$.
Here, the set $\mathcal{I}$ is constructed by sorting the projected Gaussians by depth with respect to the image plane, and $\boldsymbol{\mu}_i$ and $\Sigma_i$ are the marginalized 2D mean and variance of the $i$-th splatted Gaussian.
As discussed in \cite{kerbl20233d}, the $\alpha$-blending in \eqref{eq:gaussian-render} saturates once $\sum_{i \in \mathcal{I}} \alpha_i = 1$ and all remaining Gaussians in the ordered set $\mathcal{I}$ are not considered.
Notably, the ordering of the Gaussian splats is approximated because the rotation of each 3D Gaussian is neglected \cite{kerbl20233d}.

This splatting process may also be interpreted as marginalizing the Gaussian basis functions onto the ray creating a Gaussian Mixture Model.
The weights of each mode in the GMM correspond to the opacities computed in the alpha compositing process (Fig. \ref{fig:diagram}).
Using the 3DGS model, the moments for feature $\rho$ are computed from \eqref{eq:rendering-moments} as:
\begin{equation}
     \E[\rho^j] =  \sum_{i\in \mathcal{I}} \rho_i^j \Tilde{\alpha}_i(\boldsymbol{x}, \boldsymbol{d}) \prod_{k<i}(1-\Tilde{\alpha}_k(\boldsymbol{x}, \boldsymbol{d})).
\end{equation}

\section{Experiments} \label{sec:experiments}

This section demonstrates the utility of the proposed uncertainty quantification scheme across two case studies and validates the variance as a reasonable choice to quantify the uncertainty.
First, we demonstrate that the variance renders strongly correlate with the rendering error.
Next, we show that variance renders using our proposed approach provide a better Next-Best-View (NBV) selection criteria than existing state-of-the-art methods.
Additional implementation details and qualitative comparisons are presented in Appendices A and B, respectively.

\subsection{Error Correlation}

\begin{table*}[!t]
\footnotesize
\centering
\caption{NBV selection on the Blender dataset with a total of 12 views and the TUM dataset with a total of 300 views. Times correspond to choosing the first NBV in the experiment. Times for the random baseline are excluded since it does not consider uncertainty and relies solely on a stochastic process. Cells are highlighted with red, orange, and yellow for for the first, second, and third best performance, respectively.}
\begin{tabular}{l|ccc|c|ccc|c}
    \toprule
     \multirow{2}{*}{Method} & \multicolumn{4}{c|}{Blender} &  \multicolumn{4}{c}{TUM Dataset} \\
     & PSNR ($\uparrow$) & SSIM ($\uparrow$) & LPIPS ($\downarrow$) & Time [s] & PSNR ($\uparrow$) & SSIM ($\uparrow$) & LPIPS ($\downarrow$) & Time [s] \\
     \midrule \midrule
     Random & \cellcolor{tabthird}23.072 & \cellcolor{tabfirst}0.998 & \cellcolor{tabthird}0.162 & - &\cellcolor{tabsecond}19.175 & \cellcolor{tabfirst}0.703 & \cellcolor{tabfirst}0.382 & - \\
     ActiveNeRF \cite{pan2022activenerf} & 14.115 & \cellcolor{tabthird}0.806 & 0.280 & \cellcolor{tabthird}$1014.$ &10.834 & 0.428 & 0.613 & \cellcolor{tabthird}$1272.8$ \\
     FisherRF \cite{jiang2023fisherrf} & \cellcolor{tabsecond}23.658 & \cellcolor{tabfirst}0.998 & \cellcolor{tabsecond}0.148 & \cellcolor{tabsecond}$13.8$ &  \cellcolor{tabthird}18.978 & \cellcolor{tabsecond}0.700 & \cellcolor{tabsecond}0.388 & \cellcolor{tabsecond}$47.4$ \\
     Ours (Depth Var.) & 22.867 & \cellcolor{tabsecond}0.997 & 0.185 & \cellcolor{tabfirst}$0.6$ & 18.946 & \cellcolor{tabthird}0.697 & \cellcolor{tabsecond}0.388 & \cellcolor{tabfirst}$0.9$ \\
     Ours (Color Var.) & \cellcolor{tabfirst}23.944 & \cellcolor{tabfirst}0.998 & \cellcolor{tabfirst}0.128 & \cellcolor{tabfirst}$0.6$ &  \cellcolor{tabfirst}19.191 & \cellcolor{tabfirst}0.703 & \cellcolor{tabfirst}0.382 & \cellcolor{tabfirst}$0.9$ \\
     \bottomrule
\end{tabular}
\label{table:nbv}
\end{table*}

We determine the relationship between our proposed variance rendering approach and the rendering error for color, depth, and semantics.
A correlation between variance and error indicates that an approach correctly models uncertainty in the Gaussian splat and is able to determine regions for which the 3DGS model is a poor fit to the scene.

In this case study, we compute the Pearson correlation coefficient $\tau_p$ \cite{cohen2009pearson}, Spearman rank-order coefficient $\tau_s$ \cite{sedgwick2014spearman}, and Kendall tau coefficient $\tau_k$ \cite{sen1968estimates}.
These coefficients each model the strength and direction of correlation using different metrics, with the Pearson coefficient being the most robust to outliers.
A coefficient of 1 indicates a strong correlation, meaning regions of high rendered variance correspond to regions of high rendering error.
A coefficient of -1 indicates a strong inverse correlation, and a coefficient of 0 indicates no correlation.

We compare our proposed variance rendering approach with Bayes' Rays \cite{goli2024bayes}, FisherRF \cite{jiang2023fisherrf}, CF-NeRF \cite{yan2024cf}, and 3DGS Ensembles \cite{sunderhauf2023density}, four state-of-the-art methods that approximate model output uncertainty.
The 3DGS Ensemble approach is the 3DGS equivalent of NeRF ensembles \cite{sunderhauf2023density} and can be interpreted as a Monte Carlo Sampling-based approach for computing the integral in \eqref{eq:variance_render}.

\textbf{Color}: For color, the correlation coefficients for the Blender \cite{mildenhall2021nerf}, Mip360 \cite{barron2021mip}, and TUM \cite{sturm12iros} datasets for all methods are presented in Table \ref{table:color_var}.
As shown in Table \ref{table:color_var}, our method outperforms the FisherRF and CF-NeRF baselines in terms of correlation and runtime for every dataset and has comparable performance to the 3DGS Ensemble in terms of correlation but requires an order-of-magnitude less time to run.

\textbf{Depth}: The depth correlation coefficients are presented in Table \ref{table:depth_var}.
Here, we see the proposed method outperforms all baselines across both datasets in terms of both correlation and runtime.
Figure \ref{fig:depth_variance} shows a qualitative example of the ground truth depth error associated with a NeRF and 3DGS model and the associated uncertainty quantifications for the Bayes' Rays and FisherRF baselines and our proposed approach.
Bayes' Rays and our proposed approach estimate uncertainty using the same trained NeRF model.
Likewise, FisherRF and our proposed approach estimate uncertainty using the same trained 3DGS model.
Notably, Bayes' Rays and FisherRF overestimate uncertainty for the object while our approach estimates uncertainty over the entire scene and at the object boundaries where the depth error is highest.

\textbf{Semantics}: Table \ref{table:semantic_var} presents the semantic correlation results of the proposed approach using the Mip360 datasets.
The ground truth visual-language embeddings are computed by passing the training RGB images through the CLIP-LSeg network \cite{li2022language}.
Intuitively, conflicting visual-language embeddings will result in high variance.
We note that the rendered language features are compressed via an encoding network in \cite{zhou2024feature}, and so to compare the rendered feature vectors with the LSeg outputs, principle component analysis (PCA) is run to extract the first 128 principle components.
As there are no baselines for computing variance for language embeddings, only the results for the proposed approach are shown.
Furthermore, the ensemble method used for color and depth uncertainty quantification is infeasible for semantic uncertainty quantification due to the prohibitively expensive memory resources required to store the semantic feature vectors.
The correlation between semantic error and variance is noticeably worse than for depth and color.
We postulate that this is due to the noise inherent in the CLIP-LSeg model, where visual-language features vary for the same location depending on the viewpoint.

\begin{figure}
    \centering
    \includegraphics[width=\linewidth]{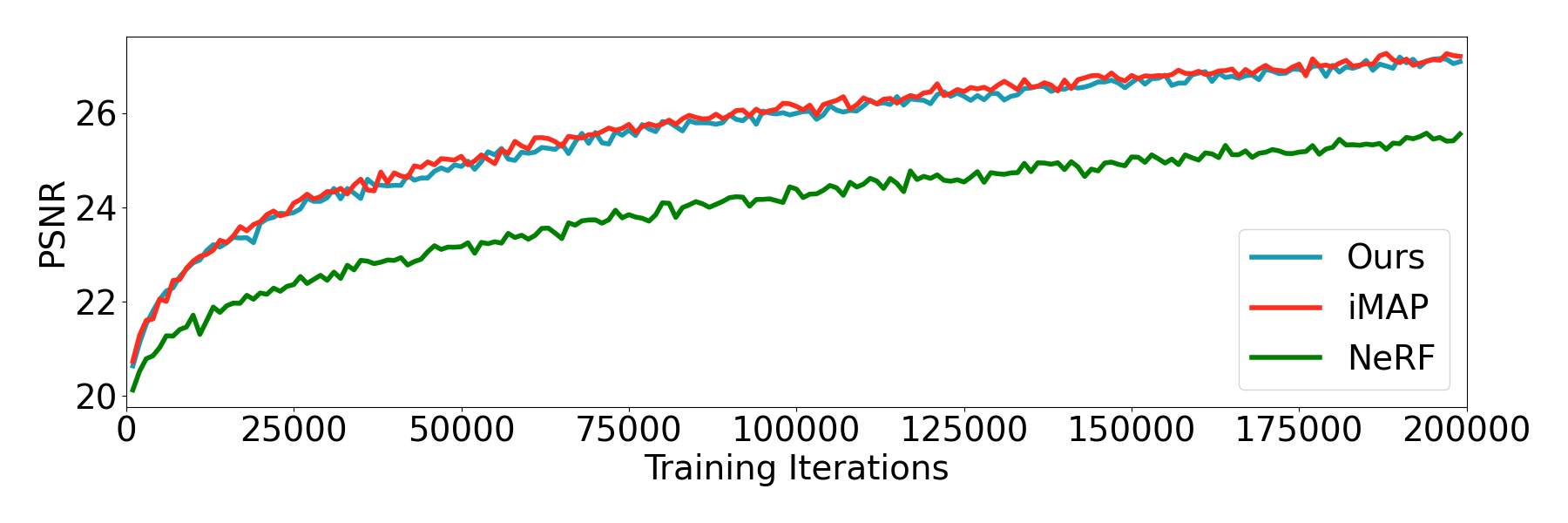}
    \caption{Rendering quality using active ray selection for the ficus scene in the Blender dataset. For the NeRF baseline, rays are selected uniformly over the entire image. iMAP \cite{sucar2021imap} and the proposed approach sample for high error and high variance regions of the image, respectfully. The proposed approach matches the performance of iMAP without needing to reference the ground-truth image.}
    \label{fig:active_ray_ficus}
\end{figure}

\textbf{Latency}: Lastly, we highlight the significantly reduced runtime of our proposed approach in comparison to the baselines as seen in Tables \ref{table:color_var} and \ref{table:depth_var}.
Our method computes variance directly from a radiance field model, in contrast to Cf-NeRF \cite{yan2024cf} and 3DGS Ensembles \cite{sunderhauf2023density}, and requires no additional post-processing, in contrast to FisherRF \cite{jiang2023fisherrf} and Bayes' Rays \cite{goli2024bayes}.


\subsection{Next-Best-View Selection}

\begin{figure}
    \centering
    \includegraphics[width=0.85\linewidth]{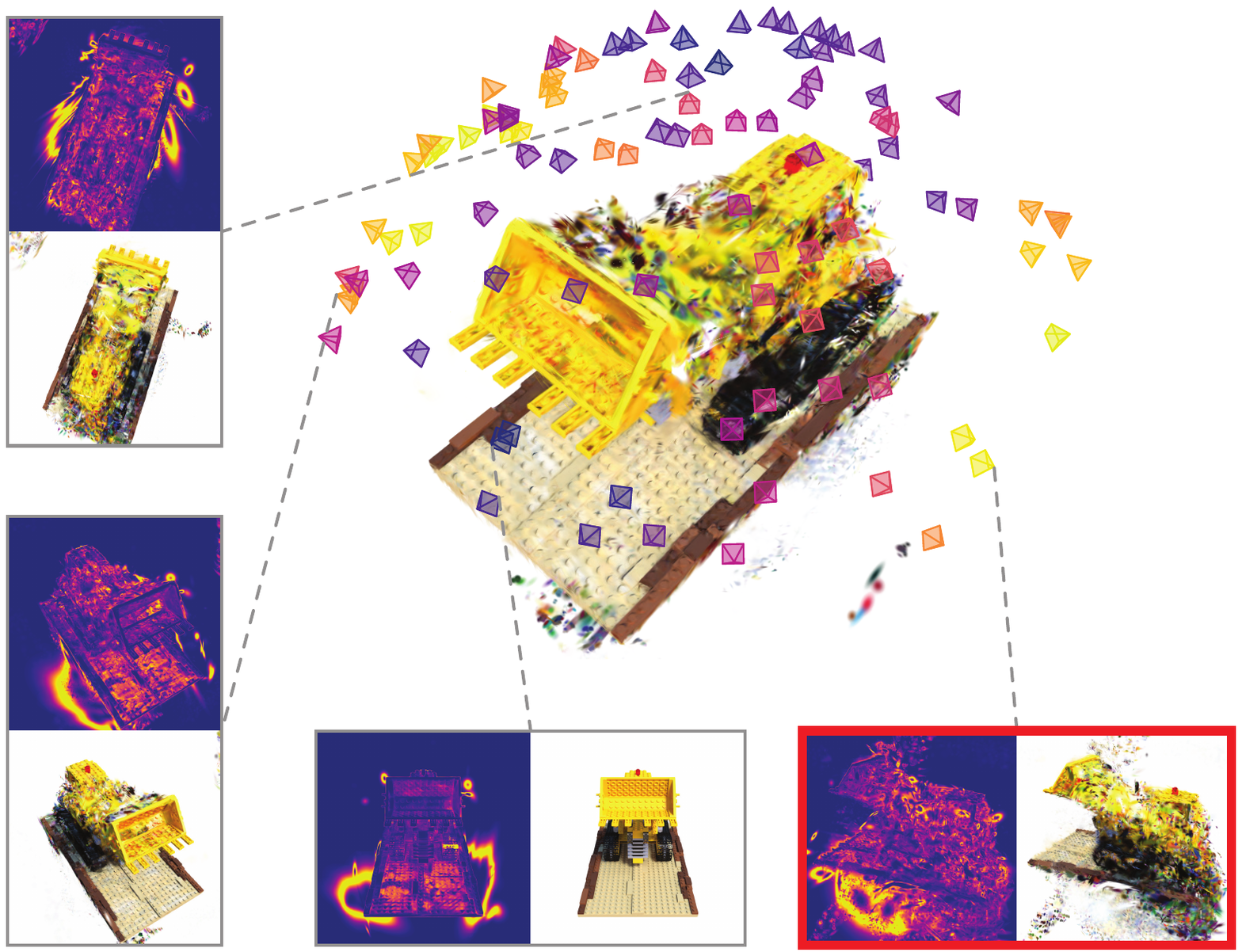}
    \caption{Given a radiance field model that is trained using only a small number views, the goal is to find the next best view, or the view that improves the radiance field model the most. The rendered variance four four views is shown. The highlighted viewpoint has the highest pixel-wise variance and is therefore chosen as the NBV.}
    \label{fig:nbv}
\end{figure}

The Next-Best-View (NBV) problem (Fig. \ref{fig:nbv}) in computer vision refers to the task of determining the optimal subsequent viewpoint that maximizes the information gained, often in the context of 3D scene reconstruction, object recognition, or inspection tasks. 
This problem is crucial in scenarios where capturing a comprehensive model of an environment or object with minimal sensing actions is desired. 
Existing methods for NBV estimate the uncertainty or entropy of the model parameters directly \cite{goli2024bayes, jiang2023fisherrf}, and then approximate model output uncertainty as a function of parameter uncertainty.
The viewpoint with the highest model parameter uncertainty is selected and added to the training set, thus improving scene completeness or reconstruction accuracy by providing more information about these parameters.

We demonstrate that computing model output uncertainty directly is a stronger selection criteria for NBV selection by comparing this approach to state-of-the-art NBV methods for radiance fields.
Following \cite{jiang2023fisherrf}, the viewpoint with the highest uncertainty is selected as the NBV.

We compare our proposed variance rendering approach with ActiveNeRF \cite{pan2022activenerf} and FisherRF \cite{jiang2023fisherrf}, two state-of-the-art methods that approximate model output uncertainty, and a random baseline. 
Notably, FisherRF \cite{jiang2023fisherrf} requires iterating over the entire training set before uncertainty is estimated for a novel viewpoint.
The Blender and TUM datasets are used for this experiment.
For the Blender dataset, the 3DGS model is initialized by uniformly sampling Gaussians within a cube with randomized RGB values and 60\% default opacities.
The model is initially trained from 9 adjacent views, and 3 new viewpoints are chosen at 1000, 3000, and 5000 iterations, respectively. 
Model performance is evaluated after 10000 training iterations.

For the TUM dataset, the model is initially trained on 150 images as the included scenes are more complex than the object-centric Blender dataset.
A new image is then added every 100 iterations until a total of 300 images are selected.
Model performance is evaluated after 20000 training iterations.
The results for both NBV experiments are presented in Table \ref{table:nbv} and assess the PSNR, SSIM, and LPIPS metrics alongside the computation time to pick an NBV.

From these results we see that using the proposed color variance for NBV selection outperforms the baselines on both datasets.
For the Blender dataset, the intricate geometric detail translates to high entropy using FisherRF \cite{jiang2023fisherrf}, resulting in a diverse set of NBV selections.
In contrast, the TUM dataset contains regions of homogeneity, such as desks and walls, which translate into low entropy using the method presented, FisherRF. 
As a result, the viewpoints selected by this method are therefore less informative for the TUM dataset as they prioritize close-up views of small objects with intricate geometry within the room that have high entropy. 
Notably, the random baseline performs surprisingly well across all experiments, with exceptional performance in the TUM dataset.

The proposed variance rendering method estimates uncertainty of the model output directly, choosing viewpoints with high variance directly without estimating model parameter uncertainty.
We propose that the reason the color variance is a better selection criteria than depth variance for the NBV task is that the evaluation criteria, PSNR, SSIM, and LPIPS, are representative of the quality of the color render.
We highlight that our approach renders variance images faster than FisherRF \cite{jiang2023fisherrf} which requires the cumulative entropy over all training images to be evaluated.


\subsection{Uncertainty-Aware Ray Sampling}
In NeRF training, rendering and optimizing over all pixels at each iteration is prohibitively expensive. 
Active sampling has emerged as an effective strategy to concentrate resources on the most informative regions, accelerating training and improving final reconstruction quality.

We propose a novel active sampling approach using the proposed variance rendering scheme in \eqref{eq:nerf_moments}.
Similar to iMAP \cite{sucar2021imap} which is a state-of-the-art approach for active ray sampling, we compute the variance for each cell of a grid spanning the image and compute the average ray variance within each cell.
In contrast, iMAP uses the average rendering error for each cell.
More training rays are then sampled from cells with higher variance.
For additional implementation details, refer to Appendix \ref{appendix:A}.

We compare the proposed approach against the original NeRF and iMAP implementations on the Blender dataset.
As seen in Figure \ref{fig:active_ray_ficus}, our proposed approach matches the state-of-the-art performance of iMAP while only requiring variance estimates along each ray rather than the rendering error.
Both methods significantly improve the final rendering quality when compared to the uniform ray sampling in the baseline NeRF implementation.
This further solidifies the correlation between the rendering error, which is used for ray selection in iMAP, and variance, used in the proposed method, and demonstrates the generality of our approach for both 3DGS and NeRF models.
Results for additional scenes are included in Appendix \ref{appendix:B} and all scenes show comparable performance between iMAP and our proposed approach.
\section{Conclusion} \label{sec:conclusion}
We present a method for rendering pixel-wise uncertainty and entropy by interpreting the rendering equation as a probabilistic process.
This enables the computation of higher-order moments and entropy for each pixel in the rendered image with minimal computational overhead.
Render speeds remain high, outperforming existing uncertainty approximation schemes for radiance field models.

Our proposed approach has a stronger correlation with rendering error than existing uncertainty quantification methods.
This indicates that the rendered variance approximates regions of the scene for which the 3DGS model is a poor fit.
Furthermore, we demonstrate that the rendered variance is a better selection criteria than existing NBV approaches.

As the pixel-wise variance and entropy are functions of the radiance field model itself, the rendered uncertainty is prone to erroneous or blurry estimates when the model has not converged.
Likewise, the pixel-wise independence approximation used in the NBV experiment does not account for the fact that a Gaussian may effect multiple pixels.
Computing the true joint probability, and resultant uncertainty, is an area of future work.
{
    \small
    \bibliographystyle{ieeenat_fullname}
    \bibliography{main}
}

\appendix
\begin{figure}
    \centering
    \includegraphics[width=\linewidth]{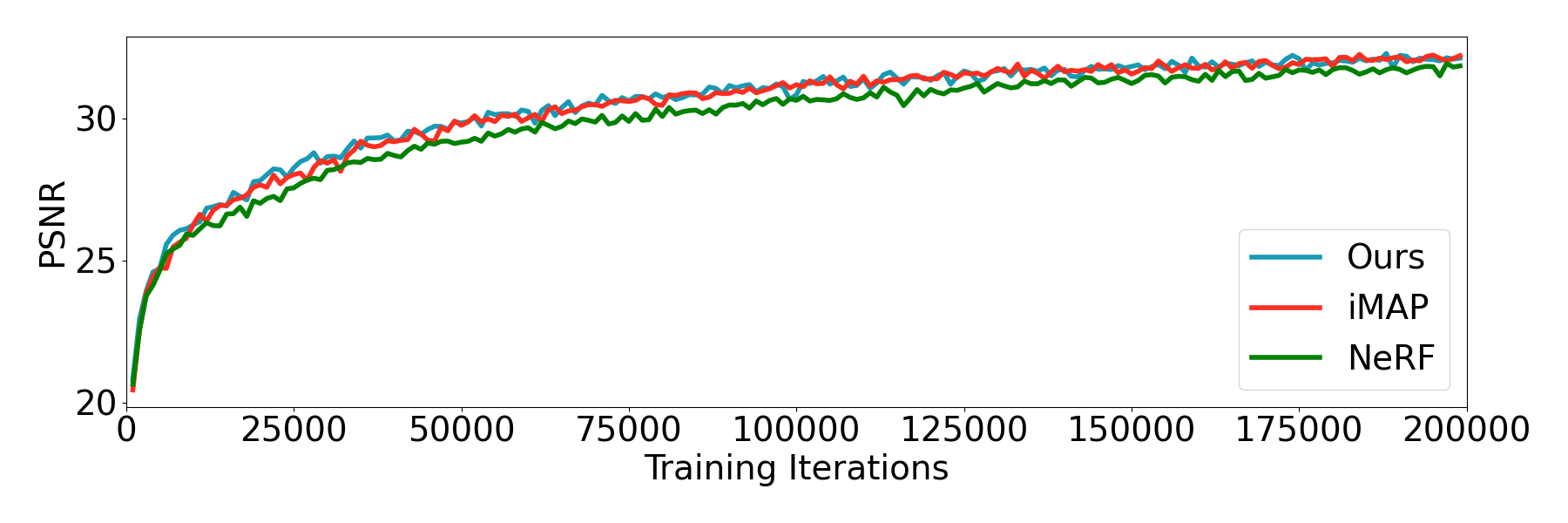}
    \caption{Rendering quality using active ray selection for the lego scene in the \textit{Blender} dataset. For the NeRF baseline, rays are selected uniformly over the entire image. iMAP \cite{sucar2021imap} and the proposed approach sample for high error and high variance regions of the image, respectfully. The proposed approach matches the performance of iMAP without needing to reference the ground-truth image. All methods have similar performance for this scene.}
    \label{fig:lego_ray}
\end{figure}

\begin{figure}
    \centering
    \includegraphics[width=\linewidth]{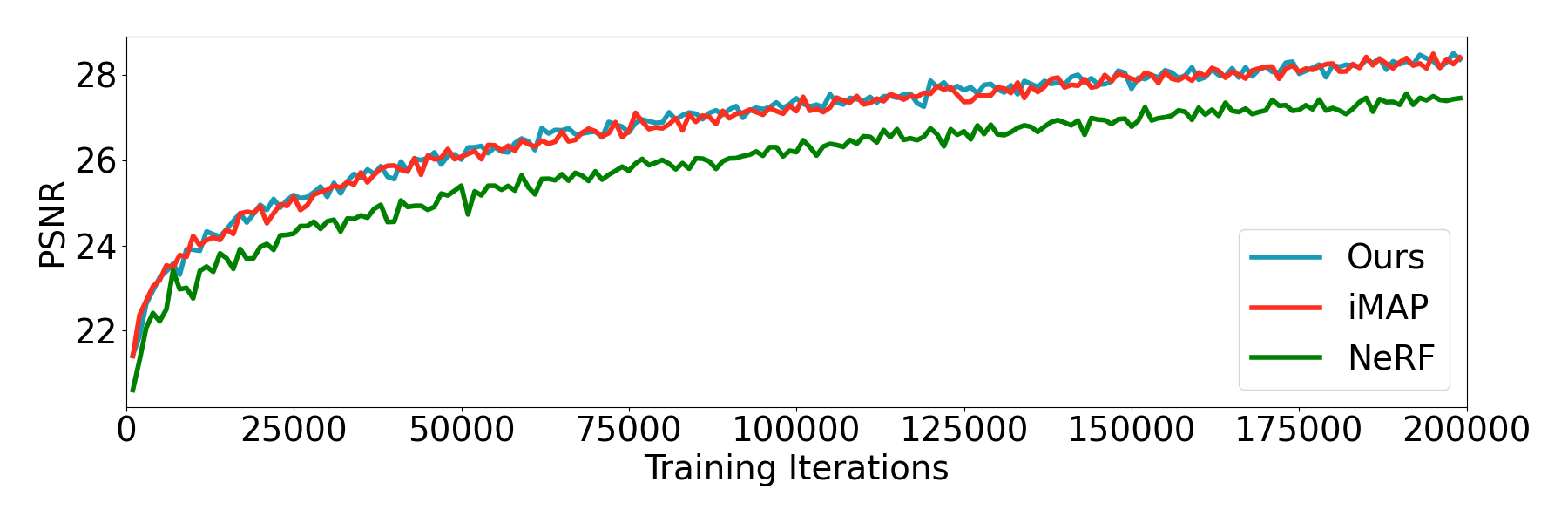}
    \caption{Rendering quality using active ray selection for the materials scene in the \textit{Blender} dataset. For the NeRF baseline, rays are selected uniformly over the entire image. iMAP \cite{sucar2021imap} and the proposed approach sample for high error and high variance regions of the image, respectfully. The proposed approach matches the performance of iMAP without needing to reference the ground-truth image. Both iMAP and our proposed approach both exceed the performance of the standard NeRF implementation.}
    \label{fig:materials_ray}
\end{figure}

\section{Implementation Details} \label{appendix:A}
\subsection{Error Correlation}
FisherRF \cite{jiang2023fisherrf}, 3DGS Ensembles \cite{sunderhauf2023density}, and the proposed approach are trained for 15000 iterations on the \textit{Blender} dataset and 30000 iterations on the \textit{Mip360} and \textit{TUM} datasets.
The CF-NeRF \cite{yan2024cf} and Bayes' Rays \cite{goli2024bayes} baselines are trained for 30000 iterations on all three datasets.

The Feature 3DGS \cite{zhou2024feature} approach is used to train the open-set visual-language semantic 3DGS model for the \textit{Mip360} dataset.
This model uses the CLIP-LSeg \cite{li2022language} visual-language model to estimate the 512-dimensional pixel-wise visual-language features for each input image, and uses the first 128 principle components to train the 3DGS model.

To compute the correlation coefficients, the 2-norm for the color and semantic rendering error images are computed along the channel dimension.

\subsection{Rendering Times}
Computation times reported in Tables 1 and 2 are the rendering times for a single uncertainty image.
For Bayes' Rays \cite{goli2024bayes}, this is the time to compute the spatial perturbations over the volumetric uncertainty field.
For FisherRF \cite{jiang2023fisherrf}, computing the uncertainty requires first iterating over all the training images, computing the Hessian of the image as a function of the image gradient with respect to the Gaussian parameters, and iteratively summing these Hessians.
We note that the rendering time for the FisherRF Gaussian rasterizer, modified from the original 3DGS implementation, is 0.3 ms, 637.3 ms, and 169.4 ms for the \textit{Blender}, \textit{Mip360}, and \textit{TUM} datasets, respectively.
Rendering times for the 3DGS Ensemble \cite{sunderhauf2023density} method are for rendering 10 images in series and computing the mean and variance of the resulting set.
The CF-NeRF \cite{yan2024cf} rendering times are computed using the implementation provided in the original paper and require first sampling from the Conditional Normalizing Flow network and then computing the mean and variance from the samples.
The rendering time for our proposed approach is the time required for the forward pass through the Gaussian rasterizer which computes the mean and variance for the color and depth.

\subsection{Next-Best-View Selection}

We conduct experiments on the \textit{Blender} dataset for the original NeRF \cite{mildenhall2021nerf} and iMAP \cite{sucar2021imap} implementations, beginning with 200 initial rays per image and then re‐sampling 1,024 rays at each iteration. 
All methods train for 200,000 iterations, with quantitative performance recorded every 1,000 iterations until convergence.
The results for the lego and materials \textit{Blender} scenes are presented in Figures \ref{fig:lego_ray} and \ref{fig:materials_ray}, respectively.
For all \textit{Blender} scenes, our proposed ray variance metric shows comparable performance to the iMAP baseline, where both exceed the standard NeRF uniform ray sampling with respect to PSNR.

\section{Additional Qualitative Results} \label{appendix:B}

\begin{figure*}[!ht]
    \centering
    \includegraphics[width=\linewidth]{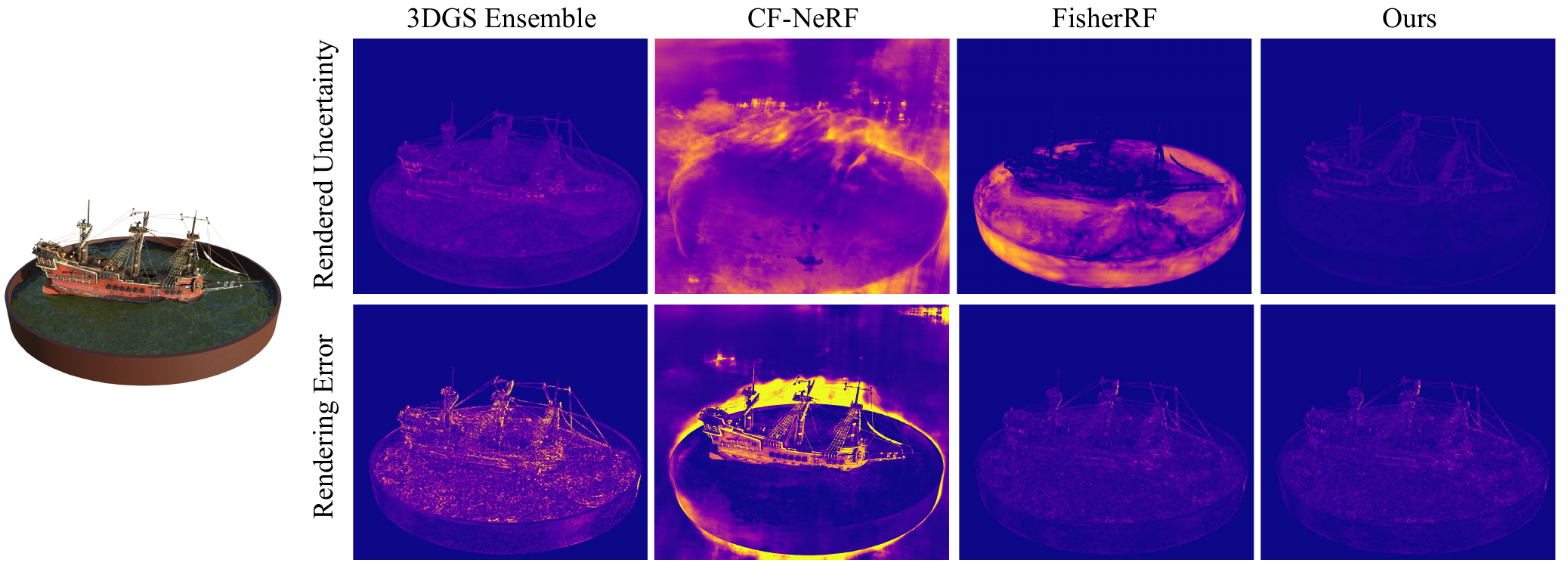}
    \caption{Qualitative color variance comparison for the ship scene from the \textit{Blender} dataset. The 3DGS Ensemble \cite{sunderhauf2023density} uses 10 trained 3DGS models to estimate the mean and variance. CF-NeRF \cite{yan2024cf} estimates uncertainty using a continuous flow network. FisherRF \cite{jiang2023fisherrf} and our proposed approach both estimate uncertainty for the same trained 3DGS model.}
    \label{fig:ship_variance}
\end{figure*}

\begin{figure*}[!ht]
    \centering
    \includegraphics[width=\linewidth]{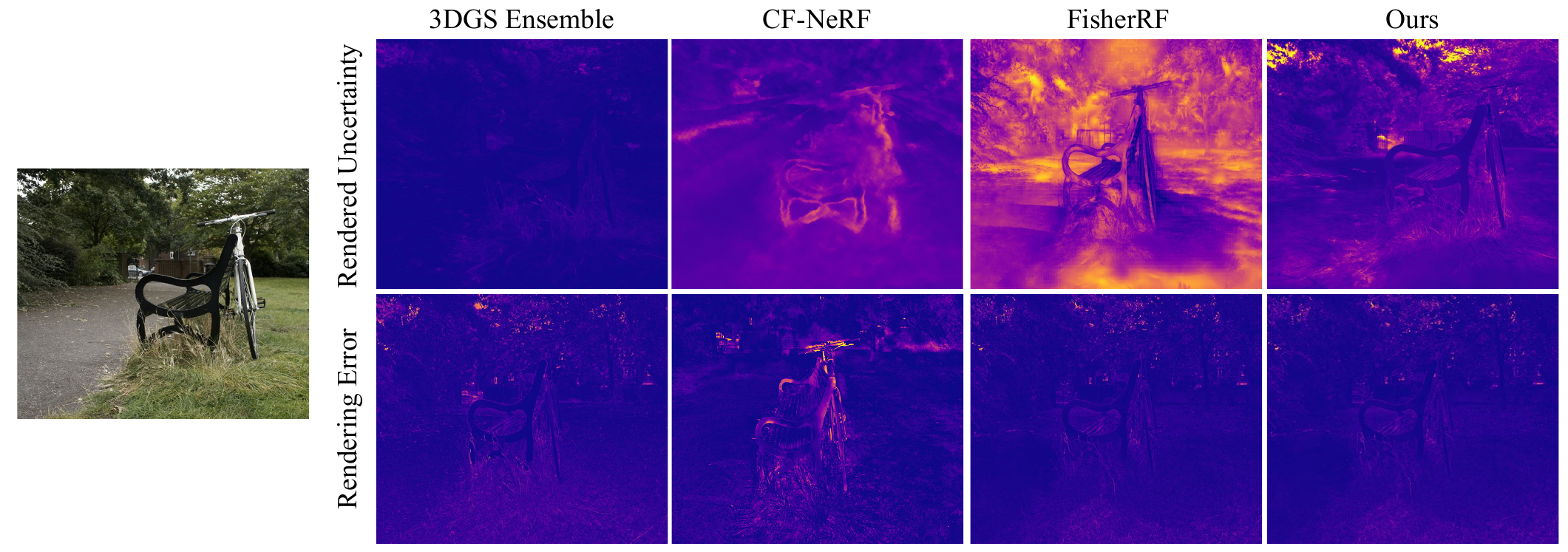}
    \caption{Qualitative color variance comparison for the bicycle scene from the \textit{Mip360} dataset. The 3DGS Ensemble \cite{sunderhauf2023density} uses 10 trained 3DGS models to estimate the mean and variance. CF-NeRF \cite{yan2024cf} estimates uncertainty using a continuous flow network. FisherRF \cite{jiang2023fisherrf} and our proposed approach both estimate uncertainty for the same trained 3DGS model.}
    \label{fig:bike_variance}
\end{figure*}

\begin{figure*}[!ht]
    \centering
    \includegraphics[width=\linewidth]{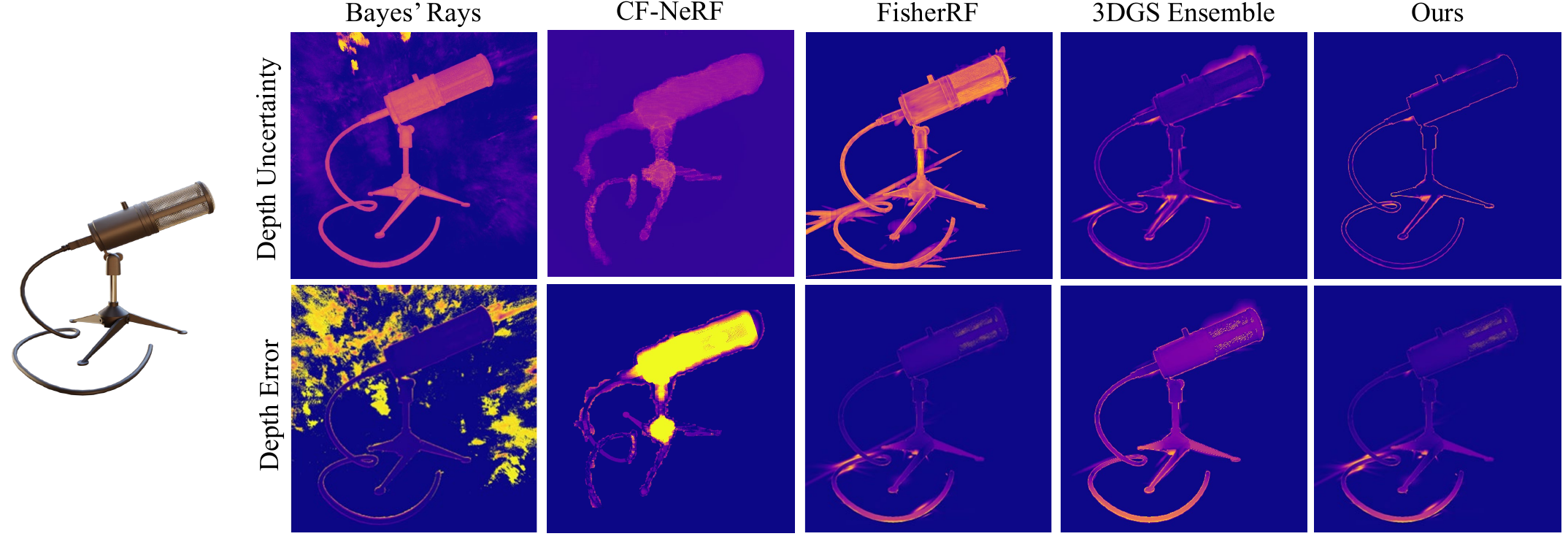}
    \caption{Qualitative depth variance comparison for the mic scene from the \textit{Blender} dataset. Bayes' Rays \cite{goli2024bayes} estimates uncertainty using a pre-trained NeRF model and a spatial uncertainty grid. The 3DGS Ensemble \cite{sunderhauf2023density} uses 10 trained 3DGS models to estimate the mean and variance. CF-NeRF \cite{yan2024cf} estimates uncertainty using a continuous flow network. FisherRF and our proposed approach both estimate uncertainty for the same trained 3DGS model.}
    \label{fig:depth_var_2}
\end{figure*}

Additional qualitative results for the proposed uncertainty quantification scheme are presented here.
Figures \ref{fig:ship_variance} and \ref{fig:bike_variance} demonstrate the qualitative correlation between rendered variance and rendering error for two scenes.
Interestingly, for our proposed approach the color variance is concentrated on the ship itself in the \textit{Blender} ship scene but not the surrounding water, whereas in the \textit{Mip360} bicycle scene the variance is concentrated on the background vegetation.

\begin{figure*}
    \centering
    \includegraphics[width=\linewidth]{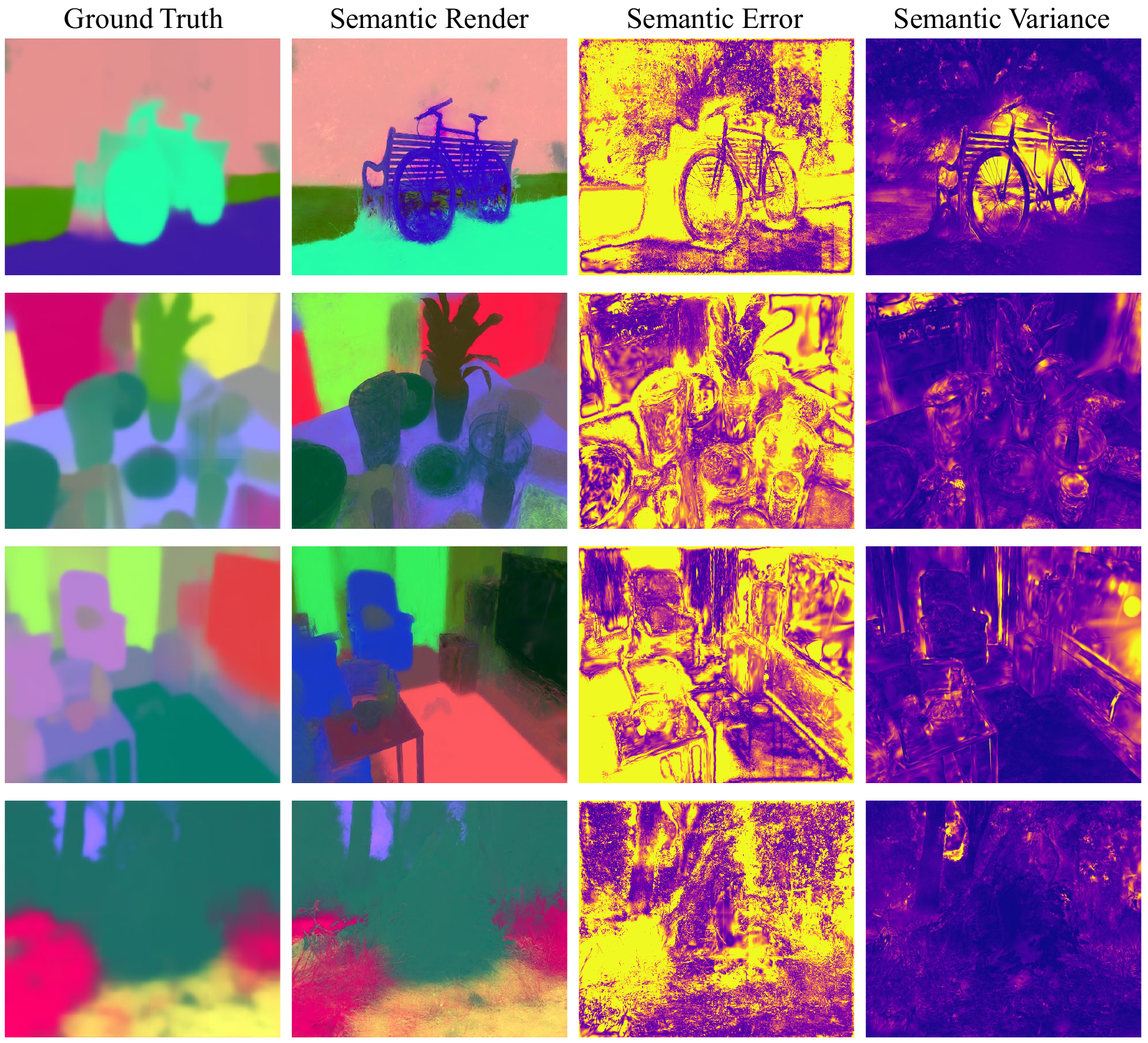}
    \caption{Qualitative semantic correlation results for several scenes from the \textit{Mip360} dataset using open-set semantics in the form of visual-language embeddings. The semantic 3DGS model is trained using Feature-3DGS \cite{zhou2024feature} by using visual-language features computed using the CLIP-LSeg network \cite{li2022language} for each input image. From the third column is seen that the semantic error is high between the CLIP-LSeg estimates and the semantic 3DGS model. We postulate this is due to the CLIP-LSeg model estimating inconsistent visual-language features between viewpoints.}
    \label{fig:open_seg}
\end{figure*}

Qualitative results for the open-dictionary semantic correlation are also presented for scenes from the \textit{Mip360} dataset in Figure \ref{fig:open_seg}.
We highlight that we use the Feature-3DGS \cite{zhou2024feature} for training the semantic 3DGS model.
Feature-3DGS passes each training image through the CLIP-LSeg \cite{li2022language} visual-language model for pixel-wise language embeddings.
Language features are visualized by computing the first three principle components, however the semantic error is computed by considering the entire 128-dimensional feature at each pixel.
From Figure \ref{fig:open_seg}, we see that there is a large discrepancy between the semantic estimates from the trained model and the language features output from the CLIP-LSeg model.
We postulate that this is the reason for the low semantic correlation score.

\begin{figure*}
    \centering
    \includegraphics[width=\linewidth]{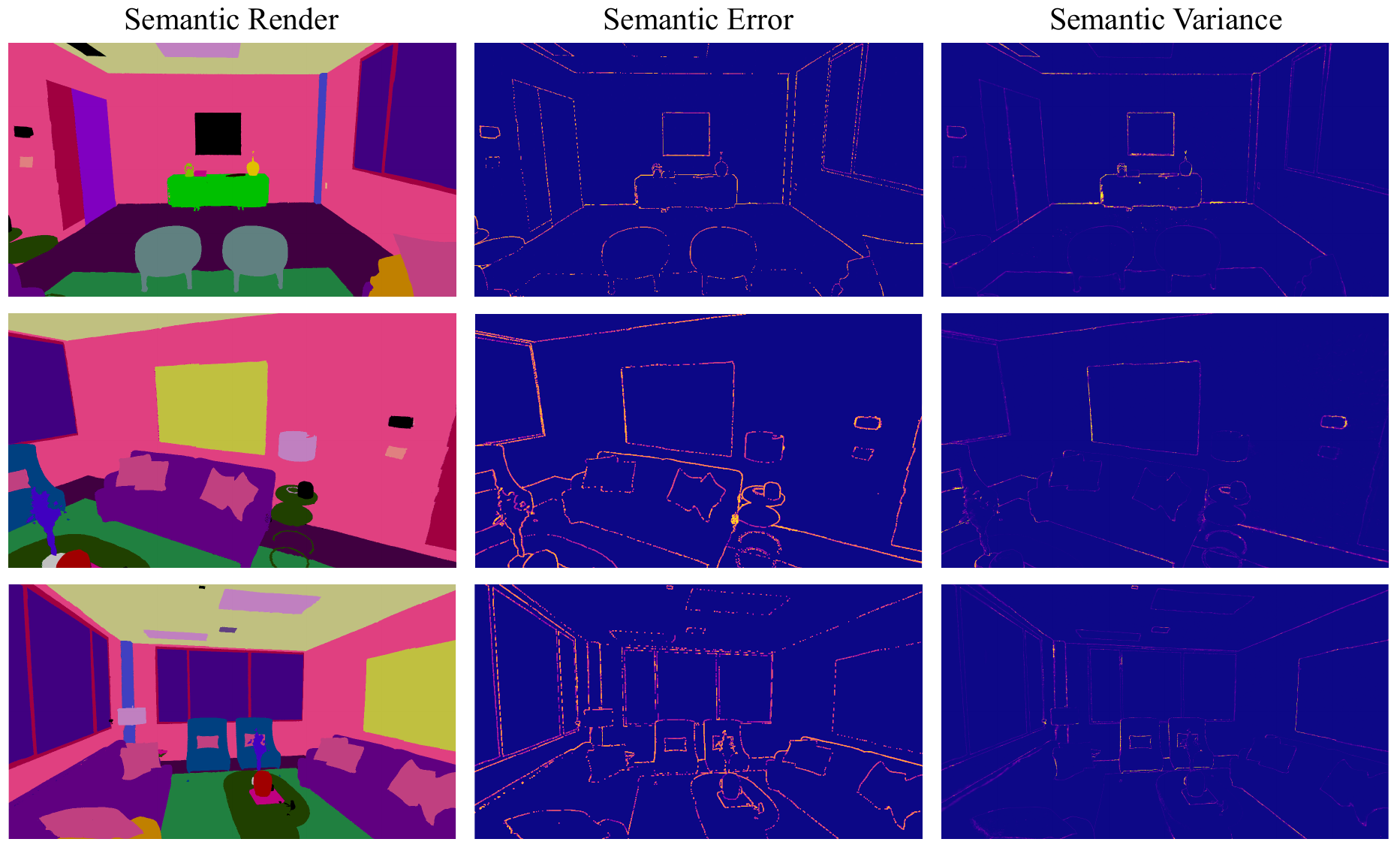}
    \caption{Qualitative semantic correlation results for a scene from the \textit{Replica} dataset using closed-set semantics. The semantic 3DGS model is trained using the SGS-SLAM \cite{li2024sgs} method. Shown in the second columns is the semantic which is concentrated on the object boundaries. This phenomenon has been observed in voxel-based closed-set semantic mapping \cite{wilson2024convbki}. The semantic variance, shown in the third column, is likewise concentrated at the object boundaries.}
    \label{fig:closed_seg}
\end{figure*}

We also provide qualitative results for the semantic correlation for a closed-set semantic 3DGS model.
The closed-set semantic 3DGS model is trained using the SGS-SLAM \cite{li2024sgs} framework for the room scene from the Replica dataset \cite{straub2019replica}, however we do not compute correlation coefficients for this model as only one scene was trained for visualization purposes.
Figure \ref{fig:closed_seg} highlights these results.
Notably, for closed-set semantics, the semantic error is concentrated at the boundaries of objects, a phenomenon also observed in voxel-based uncertainty quantification methods \cite{wilson2024convbki}.

\end{document}